\documentclass[10pt,twocolumn,letterpaper]{article}

\usepackage{iccv}
\usepackage{times}
\usepackage{epsfig}
\usepackage{graphicx}
\usepackage{amsmath}
\usepackage{amssymb}

\usepackage{algorithm}
\usepackage{algorithmic}
\usepackage{bm}
\usepackage{mathtools}
\usepackage{makecell}
\usepackage{enumitem}
\usepackage{caption}
\usepackage{lipsum}
\usepackage{cuted}
\usepackage{siunitx}
\usepackage{booktabs}

\usepackage{subcaption}
\usepackage{multirow}
\makeatletter
\DeclareRobustCommand\onedot{\futurelet\@let@token\@onedot}
\def\@onedot{\ifx\@let@token.\else.\null\fi\xspace}

\def\eg{\emph{e.g}\onedot} 
\def\ie{\emph{i.e}\onedot} 
 	
\def\etc{\emph{etc}\onedot} \def\vs{\emph{vs}\onedot}

\makeatother

\usepackage[pagebackref=true,breaklinks=true,letterpaper=true,colorlinks,bookmarks=false]{hyperref}

\iccvfinalcopy 

\def\iccvPaperID{6957} 

\ificcvfinal\fi

\begin{document}

\title{ResRep: Lossless CNN Pruning via Decoupling Remembering and Forgetting}

\author{\textbf{Xiaohan Ding \textsuperscript{1,2} \quad Tianxiang Hao \textsuperscript{1,2} \quad Jianchao Tan \textsuperscript{3} \quad Ji Liu \textsuperscript{3}} \\
	\textbf{Jungong Han \textsuperscript{4}  \quad Yuchen Guo \textsuperscript{1} \quad Guiguang Ding \textsuperscript{1,2}} \thanks{This work was supported by the National Natural Science Foundation of China (No. 61925107, U1936202, 61971260), the National Key R$\&$D Program of China (No. 2020AAA0105500). Xiaohan Ding is funded by the Baidu Scholarship Program 2019. Correspondence to: Yuchen Guo, Guiguang Ding.}  \\
	\textsuperscript{1} Beijing National Research Center for Information Science and Technology (BNRist) \\
	\textsuperscript{2} School of Software, Tsinghua University, Beijing, China \\
	\textsuperscript{3} AI platform department, Seattle AI lab, and FeDA lab, Kwai Inc. \\
	\textsuperscript{4} Computer Science Department, Aberystwyth University, SY23 3FL, UK \\
	\tt\small dxh17@mails.tsinghua.edu.cn \quad beyondhtx,tanjianchaoustc,ji.liu.uwisc@gmail.com \\
	\tt\small jungonghan77,yuchen.w.guo@gmail.com \quad dinggg@tsinghua.edu.cn \\
}

\maketitle
\ificcvfinal\thispagestyle{empty}\fi

\begin{abstract}
   We propose ResRep, a novel method for lossless channel pruning (a.k.a. filter pruning), which slims down a CNN by reducing the width (number of output channels) of convolutional layers. Inspired by the neurobiology research about the independence of remembering and forgetting, we propose to re-parameterize a CNN into the remembering parts and forgetting parts, where the former learn to maintain the performance and the latter learn to prune. Via training with regular SGD on the former but a novel update rule with penalty gradients on the latter, we realize structured sparsity. Then we equivalently merge the remembering and forgetting parts into the original architecture with narrower layers. In this sense, ResRep can be viewed as a successful application of Structural Re-parameterization. Such a methodology distinguishes ResRep from the traditional learning-based pruning paradigm that applies a penalty on parameters to produce sparsity, which may suppress the parameters essential for the remembering. ResRep slims down a standard ResNet-50 with 76.15\% accuracy on ImageNet to a narrower one with only 45\% FLOPs and no accuracy drop, which is the first to achieve lossless pruning with such a high compression ratio. The code and models are at \url{https://github.com/DingXiaoH/ResRep}.
\end{abstract}

\section{Introduction}

The mainstream techniques to compress and accelerate convolutional neural network (CNN) include sparsification \cite{GSM,guo2016dynamic,han2015learning}, channel pruning \cite{FPGM,he2017channel,HRank}, quantization \cite{ba2014deep,DBLP:conf/nips/BannerNS19,DBLP:conf/eccv/LiuWLYLC18,DBLP:conf/nips/ZhaoGBM019}, knowledge distillation \cite{hinton2015distilling,DBLP:conf/cvpr/JiaoWJSLH19,DBLP:conf/cvpr/LiuCLQLW19,DBLP:conf/cvpr/VongkulbhisalVS19}, \etc. Channel pruning \cite{he2017channel} (a.k.a. filter pruning \cite{li2016pruning} or network slimming~\cite{liu2017learning}) reduces the width (\ie, number of output channels) of convolutional layers to effectively reduce the number of floating-point operations (FLOPs) and memory footprint, which is complementary to the other model compression methods as it produces a thinner model of the original architecture with no custom structures or operations.

\begin{figure*}
	\begin{center}
		\includegraphics[width=0.9\linewidth]{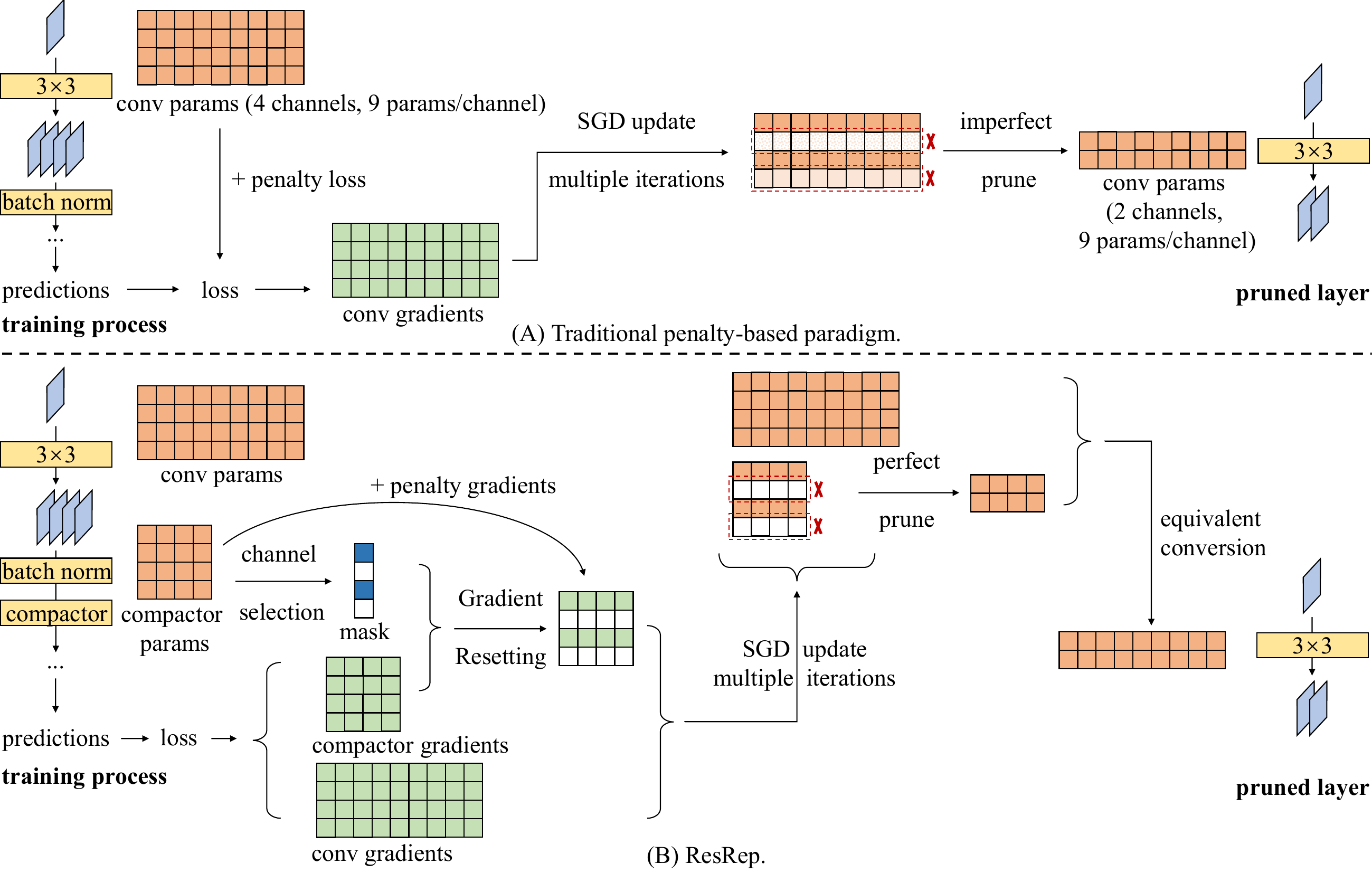}
		\vspace{-0.1in}
		\caption{Traditional penalty-based pruning \vs ResRep. We prune a $3\times3$ layer with one input channel and four output channels for illustration. For the ease of visualization, we ravel the kernel $\bm{K}\in\mathbb{R}^{4\times1\times3\times3}$ into a matrix $\bm{W}\in\mathbb{R}^{4\times9}$. \textbf{A)} To prune some channels of $\bm{K}$ (\ie, rows of $\bm{W}$), we add a penalty loss on the kernel to the original loss, so that the gradients will make some rows smaller in magnitude, but not small enough to realize perfect pruning. \textbf{B)} ResRep constructs a compactor with kernel matrix $\bm{Q}\in\mathbb{R}^{4\times4}$. Driven by the penalty gradients, the compactor selects some of its channels and generates a binary mask, which resets some of the original gradients of $\bm{Q}$ to zero. After multiple iterations, those compactor channels with reset gradients become infinitely close to zero, which enables perfect pruning. Finally, the conv-BN-compactor sequence is equivalently converted into a regular conv layer with two channels. Blank rectangles indicate zero values.}
		\label{fig-intro}
	\end{center}
	\vspace{-0.2in}
\end{figure*}

However, as CNN's representational capacity depends on the width of conv layers, it is difficult to reduce the width without performance drops. On practical CNN architectures like ResNet-50 \cite{he2016deep} and large-scale datasets like ImageNet \cite{deng2009imagenet}, lossless pruning with high compression ratio has long been considered challenging. For reasonable trade-off between compression ratio and performance, a typical paradigm (Fig. \ref{fig-intro}.A) \cite{alvarez2016learning,DBLP:conf/nips/AlvarezS16,ding2018auto,lin2019towards,liu2015sparse,wang2018structured,wen2016learning} trains the model with magnitude-related penalty loss (\eg, group Lasso \cite{roth2008group,Simon2013ASL}) on the conv kernels to produce \textit{structured sparsity}, which means all the parameters of some channels become small in magnitude. Ideally, if the parameters of pruned channels are small enough, the pruned model may deliver the same performance as before (\ie, after training but before pruning), which we refer to as \textit{perfect pruning}.

Since both the training and pruning may degrade the performance, we evaluate a training-based pruning method from two aspects. \textbf{1) Resistance}. The training phase tends to decrease the accuracy (referred to as \textit{training-caused damage}) as it introduces some desired properties such as structured sparsity into the model, which may be harmful because the objective of optimization is changed and the parameters are deviated from the optima. We say a model has high resistance if the performance maintains high during training. \textbf{2) Prunability}. When we prune the model into a smaller one after training, the properties obtained (\eg, many channels being close to zero) will reduce the \textit{pruning-caused damage}. If the model endures a high pruning ratio with low performance drop, we say it has high prunability.

We desire both high resistance and prunability, but the traditional penalty-based paradigm naturally suffers from a resistance-prunability trade-off. For example, a strong group Lasso achieves high sparsity with great training-caused damage, while a weak penalty maintains the performance but results in low sparsity, hence great pruning-caused damage. Sect. \ref{sect-gr} presents detailed analysis.

In this paper, we propose ResRep to address the above problem, which is inspired by the neurobiology research on remembering and forgetting. \textbf{1)} Remembering requires the brain to potentiate some synapses but depotentiate the others, which resembles the training of CNN that makes some parameters large and some small. \textbf{2)} Synapse elimination via shrinkage or loss of spines is one of the classical forgetting mechanisms~\cite{RICHARDS20171071} as a key process to improve efficiency in both energy and space for biological neural network, which resembles pruning. Neurobiology research reveals that remembering and forgetting are independently controlled by Rutabaga adenylyl cyclase-mediated memory formation mechanism and Rac-regulated spine shrinkage mechanism, respectively~\cite{Dong2016InabilityTA,HayashiTakagi2015LabellingAO,Shuai2010ForgettingIR}, indicating it is more reasonable to learn and prune by two decoupled modules.


Inspired by such independence, we propose to decouple the ``remembering'' and ``forgetting'', which are coupled in the traditional paradigm as the conv parameters are involved in both the ``remembering'' (objective function) and ``forgetting'' (penalty loss) in order for them to achieve a trade-off. That is, the traditional methods force every channel to ``forget'', and remove the channels that ``forgot the most''. In contrast, we first re-parameterize the original model into ``remembering parts'' and ``forgetting parts'', then apply ``remembering learning'' (\ie, regular SGD with the original objective) on the former to maintain the ``memory'' (original performance), and ``forgetting learning'' on the latter to ``eliminate synapses'' (zero out channels).

ResRep comprises two key components: Convolutional Re-parameterization (\textit{Rep}, the methodology of decoupling and the corresponding equivalent conversion) and Gradient Resetting (\textit{Res}, the updage rule for ``forgetting''). Specifically, we insert a \textit{compactor}, which is a $1\times1$ conv, after the original conv layer we desire to prune. During training, we add penalty gradients to \textit{only the compactors}, select some compactor channels and zero out their gradients derived from the objective function. Such a training process makes some channels of compactors very close to zero, which are removed with no pruning-caused damage. Then we \textit{equivalently} convert the compactor together with the preceding conv into a single conv with fewer channels through some linear transformations (Eq. \ref{eq-k-prime}, \ref{eq-b-prime}). Note that this method readily generalizes to the common case where the original conv is followed by batch normalization (BN) \cite{ioffe2015batch}. In this case, we append the compactor after BN, and convert the conv-BN-compactor sequence after training by first equivalently fusing the conv-BN into a conv with bias (Eq. \ref{eq-fusebn}). Eventually, the resultant model has the same architecture as the original (\ie, no compactors) but narrower layers (Fig. \ref{fig-intro}.B). As the equivalent conversion from the training-time model into the pruned model relies on the equivalent conversion of the parameters, ResRep can be viewed as an application of Structural Re-parameterization \cite{ding2019acnet,ding2021repvgg,ding2021repmlp,ding2021diverse}. The other Structural Re-parameterization works improve the VGG-like architectures \cite{ding2021repvgg}, basic conv layers \cite{ding2019acnet,ding2021diverse} or an MLP-style building block \cite{ding2021repmlp} with different well-designed structures but all via an ordinary training process. In contrast, ResRep not only constructs extra structures (\ie, compactors) that can be equivalently converted back (Rep) but also uses a custom training strategy (Res). As will be shown in Sect. \ref{sect-ablation}, Rep and Res are both essential: Rep constructs some structures for Res to apply on without losing the original information; Res zeros out some channels so that Rep can make the resultant conv layer narrower. Notably, ResRep can also prune a fully-connected layer because it is equivalent to a $1\times1$ conv \cite{long2015fully}.

\begin{algorithm}[!h]
	\vskip -0.03in
	\caption{Pipeline of ResRep channel pruning.}
	\label{alg1}
	\begin{algorithmic}[1]
		\STATE {\bfseries Input: well-trained model $\mathcal{W}$}
		\STATE Construct the re-parameterized model $\mathcal{\hat{W}}$ with compactors. Initialize the compactors as identity matrices and the other parts with the original parameters of $\mathcal{W}$
		\FOR{$i=0$ to max training iterations}
		\STATE Forward a batch through $\mathcal{\hat{W}}$, compute the loss using the original objective function, derive the gradients
		\STATE Apply Gradient Resetting on the gradients of compactors only (Eq. \ref{eq-grad-lasso-W-m})
		\STATE Update $\mathcal{\hat{W}}$ with the reset gradients of compactors and the original gradients of the other parameters
		\ENDFOR
		\STATE Delete the rows of compactors which are close to zero (\eg, with norm $\leq 10^{-5}$) in $\mathcal{\hat{W}}$. Equivalently convert the parameters of $\mathcal{\hat{W}}$ into $\mathcal{W^{\prime}}$ (Eq. \ref{eq-fusebn}, \ref{eq-k-prime}, \ref{eq-pr-cur-layer}). Now $\mathcal{W^{\prime}}$ has the same architecture as $\mathcal{W}$ but narrower layers
		\STATE {\bfseries Output: pruned model $\mathcal{W^{\prime}}$}
	\end{algorithmic}
\end{algorithm}

ResRep features: \textbf{1)} High resistance. To maintain the performance, ResRep does not change the loss function, update rule or any training hyper-parameters of the original model (\ie, the conv-BN parts). \textbf{2)} High prunability. The compactors are driven by the penalty gradients to make many channels small enough to realize perfect pruning, even with a mild penalty strength. \textbf{3)} Given the required global reduction ratio of FLOPs, ResRep automatically finds the appropriate eventual width of each layer with no prior knowledge, making it a powerful tool for CNN structure optimization. \textbf{4)} End-to-end training and easy implementation (Alg. \ref{alg1}). We summarize our contributions as:
\begin{itemize}[noitemsep,nolistsep,topsep=0pt,parsep=0pt,partopsep=0pt,leftmargin=*]
	\item Inspired by the neurobiology research, we proposed to decouple ``remembering'' and ``forgetting'' for pruning.
	\item We proposed two techniques, Rep and Res, to achieve high resistance and prunability. They can be used separately and the combination delivers the best performance.
	\item We achieved state-of-the-art results on common benchmark models, including real lossless pruning on ResNet-50 on ImageNet with a pruning ratio of $54.5\%$.
\end{itemize}

\section{Related Work}

Pruning may refer to removing any parameters or structures from a network. Unstructured pruning \cite{GSM,guo2016dynamic,han2015deep,han2015learning} can reduce the number of non-zero parameters but cannot realize speedup on common computing frameworks. Structured pruning removes some whole structures (\eg, neurons of fully-connected layers, 2D kernels, and channels), which is more friendly to hardware \cite{liu2020autocompress,wen2016learning}. Channel pruning is especially practical as it reduces not only the model size, the actual computations, but also the memory footprint. Pruning is related to the lottery ticket hypothesis \cite{frankle2018lottery}. For example, one may use ResRep to find the ``winning'' channels before training. Except for generic model pruning, pruning in specific contexts (\eg, the limited-data scenario \cite{luo2020neural}) has also attracted much attention.

Most of the channel pruning methods can be categorized into two families. \textbf{Pruning-then-finetuning} methods identify and prune the unimportant channels from a well-trained model by some measurements \cite{abbasi2017structural,hu2016network,li2016pruning,luo2017thinet,molchanov2019importance,molchanov2016pruning,polyak2015channel,yu2018nisp}, which may cause significant accuracy drop, and finetune it afterwards. Some methods repeat pruning-finetuning iterations to measure the importance and prune progressively \cite{ding2019approximated,lin2018accelerating}. A major drawback is that the pruned models can be easily trapped into bad local minima, and sometimes cannot even reach a similar level of accuracy with a counterpart of the same structure trained from scratch \cite{liu2018rethinking}. This discovery highlights the significance of perfect pruning, which eliminates the need for finetuning. In this family, PCAS \cite{yamamoto2018pcas} is the most related to ResRep, which identifies the unimportant channels by training attention modules appended after conv layers. Unlike ResRep, PCAS performs imperfect pruning and requires finetuning after removing the unimportant channels. Moreover, PCAS discards the attention modules after training, which results in more structural damage, while ResRep uses a \textit{mathematically equivalent} transformation to obtain the final model structure without any performance drop. \textbf{Learning-based pruning} methods utilize a custom learning process to reduce the pruning-caused damage. Apart from the above-mentioned penalty-based paradigm to zero out some of the channels \cite{alvarez2016learning,DBLP:conf/nips/AlvarezS16,ding2018auto,lin2019towards,liu2015sparse,wang2018structured,wen2016learning}, some other methods prune via making some filters identical \cite{CSGD-journal}, meta-learning \cite{Metapruning}, adversarial learning \cite{GAL}, \etc.


\section{ResRep for Lossless Channel Pruning}

\subsection{Formulation and Background}
We first introduce the formulation of conv and channel pruning. Let $D$ and $C$ be the output and input channels, $K$ be the kernel size, $\bm{K}\in\mathbb{R}^{D\times C\times K\times K}$ be the kernel parameter tensor, $\bm{b}\in\mathbb{R}^D$ be the optional bias, $I\in\mathbb{R}^{N\times C\times H\times W}$ and $\bm{O}\in\mathbb{R}^{N\times D\times H^\prime \times W^\prime}$ be the input and output, $\circledast$ be the convolution operator, and $B$ be the broadcast function which replicates $\bm{b}$ into $N\times D\times H^\prime \times W^\prime$, we have
\begin{equation}\label{eq-conv-bias}
\bm{O} = \bm{I} \circledast \bm{K} + B(\bm{b}) \,.
\end{equation}

For a conv layer with no bias term but a following batch normalization (BN) \cite{ioffe2015batch} layer with mean $\bm{\mu}$, standard deviation $\bm{\sigma}$, scaling factor $\bm{\gamma}$ and bias $\bm{\beta}\in\mathbb{R}^D$, we have
\begin{equation}\label{eq-conv-bn}
\bm{O}_{:,j,:,:} = ((\bm{I} \circledast \bm{K})_{:,j,:,:} - \mu_{j})\frac{\gamma_{j}}{\sigma_{j}} + \beta_j \,, \forall 1\leq j \leq D \,.
\end{equation}

Let $i$ be the index of conv layer. To prune conv $i$, we obtain the index set of pruned channels $\mathcal{P}^{(i)} \subset \{1, 2, \dots, D\}$ according to some rules, then its complementary set $\mathcal{S}^{(i)} = \{1, 2, \dots, D\} \setminus \mathcal{P}^{(i)}$ for the index set of channels which survive. The pruning operation preserves the $\mathcal{S}^{(i)}$ output channels of conv $i$ and the corresponding input channels of the succeeding layer (conv $i+1$), and discard the others. The corresponding entries in the bias or following BN, if any, should be discarded as well. The obtained kernels are
\begin{equation}\label{eq-pr-cur-layer}
\bm{K}^{(i)\prime} = \bm{K}^{(i)}_{\mathcal{S}^{(i)},:,:,:} \,, \quad \bm{K}^{(i+1)\prime} = \bm{K}^{(i+1)}_{:,\mathcal{S}^{(i)},:,:} \,.
\end{equation}

\subsection{Convolutional Re-parameterization}\label{cr}

For every conv layer together with the following BN (if any) we desire to prune, which are referred to as the \textit{target layers}, we append a compactor ($1\times1$ conv) with kernel $\bm{Q}\in\mathbb{R}^{D\times D}$. Given a well-trained model $\mathcal{W}$, we construct a re-parameterized model $\mathcal{\hat{W}}$ by initializing the conv-BN as the original weights of $\mathcal{W}$ and $\bm{Q}$ as an identity matrix, so that the re-parameterized model produces the identical outputs as the original. Note that if the target layer has no following BN, the ``BN'' in our notation can be safely viewed as a bias. After training with Gradient Resetting, which will be described in detail in Sect. \ref{sect-gr}, we prune the resulting close-to-zero channels of compactors and convert the model into $\mathcal{W}^\prime$, which has the same architecture as $\mathcal{W}$ but narrower layers. Concretely, for a specific compactor with kernel $\bm{Q}$, we prune the channels with norm smaller than a threshold $\epsilon$. Formally, we obtain the to-be-pruned set by $\mathcal{P} = \{j \,|\, ||\bm{Q}_{j,:}||_2 < \epsilon\}$, or the surviving set $\mathcal{S} = \{j \,|\, ||\bm{Q}_{j,:}||_2 \geq \epsilon\}$. Similar to Eq. \ref{eq-pr-cur-layer}, we prune $\bm{Q}$ by $\bm{Q}^\prime = \bm{Q}_{\mathcal{S},:}$. In our experiments, we use $\epsilon = 10^{-5}$, which is found to be small enough to realize perfect pruning.\footnote{Gradient Resetting will make some channels infinitely close to zero, as shown in Fig. \ref{fig-comparison}, so setting $\epsilon = 10^{-5}$ or $\epsilon = 10^{-9}$ makes no difference.} After pruning, the compactor has fewer rows than columns, \ie, $\bm{Q}^\prime \in \mathbb{R}^{D^\prime \times D}, D^\prime = |\mathcal{S}|$. To convert $\mathcal{\hat{W}}$ into $\mathcal{W}^\prime$, we seek to convert every conv-BN-compactor sequence into a conv layer with $\bm{K}^\prime\in\mathbb{R}^{D^\prime \times C \times K\times K}$ and bias $\bm{b}^\prime \in \mathbb{R}^{D^\prime}$.

Firstly, we equivalently fuse a conv-BN into a conv for inference, which produces the identical outputs as the original. With $\bm{K}, \bm{\mu}, \bm{\sigma}, \bm{\gamma}, \bm{\beta}$ of a conv-BN, we construct a new conv with kernel $\bar{\bm{K}}$ and bias $\bar{\bm{b}}$ as follows. For $1\leq j \leq D$,
\begin{equation}\label{eq-fusebn}
\bar{\bm{K}}_{j,:,:,:} = \frac{\gamma_j}{\sigma_j}\bm{K}_{j,:,:,:} \,, \quad \bar{b}_j = -  \frac{\mu_j\gamma_j}{\sigma_j} + \beta_j \,.
\end{equation}
Given Eq. \ref{eq-conv-bias}, \ref{eq-conv-bn} and the homogeneity of conv, we can verify
\begin{equation}\label{eq-bn-fused}
((\bm{I} \circledast \bm{K})_{:,j,:,:} - \mu_{j})\frac{\gamma_{j}}{\sigma_{j}} + \beta_j = (\bm{I} \circledast \bar{\bm{K}} + B(\bar{\bm{b}}))_{:,j,:,:}  \,.
\end{equation}
Then we seek for the formula to construct $\bm{K}^\prime$ and $\bm{b}^\prime$ so that
\begin{equation}
(\bm{I} \circledast \bar{\bm{K}} + B(\bar{\bm{b}})) \circledast \bm{Q}^\prime = \bm{I} \circledast \bm{K}^\prime + B(\bm{b}^\prime) \,.
\end{equation}
With the additivity of convolution, we arrive at 
\begin{equation}\label{eq-expanded}
\bm{I} \circledast \bar{\bm{K}} \circledast \bm{Q}^\prime + B(\bar{\bm{b}})\circledast \bm{Q}^\prime  = \bm{I} \circledast \bm{K}^\prime + B(\bm{b}^\prime) \,.
\end{equation}

We note that very channel of $B(\bar{\bm{b}})$ is a constant matrix, thus every channel of $B(\bar{\bm{b}})\circledast \bm{Q}^\prime$ is also a constant matrix. As the $1\times1$ conv with $\bm{Q}^\prime$ on $\bm{I} \circledast \bar{\bm{K}}$ only performs cross-channel recombination, we can merge $\bm{Q}^\prime$ into $\bar{\bm{K}}$ by recombining the entries in $\bar{\bm{K}}$. Let $T$ be the transpose function (\eg, $T(\bar{\bm{K}})$ is $C\times D\times K\times K$), we present the formulas to construct $\bm{K}^\prime$ and $\bm{b}^\prime$, which can be easily verified.
\begin{equation}\label{eq-k-prime}
\bm{K}^\prime = T(T(\bar{\bm{K}}) \circledast \bm{Q}^\prime) \,,
\end{equation}
\begin{equation}\label{eq-b-prime}
b^\prime_j = \bar{\bm{b}} \cdot \bm{Q}^\prime_{j,:} \,, \quad \forall 1\leq j \leq D^\prime \,.
\end{equation}

In practice, we convert and save the weights of the trained re-parameterized model, construct a model with the original architecture but narrower layers without BN, and use the saved weights for testing and deployment.

\subsection{Gradient Resetting}\label{sect-gr}

We describe how to produce structured sparsity in compactors while maintaining the accuracy, beginning by discussing the traditional usage of penalty on a specific kernel $\bm{K}$ to make the magnitude of some channels smaller, \ie, $||\bm{K}_{\mathcal{P},:,:,:}||\to 0$. Let $\bm{\Theta}$ be the universal set of parameters, $X, Y$ be the data examples and labels, $L_{\text{perf}}(X, Y, \bm{\Theta})$ be the performance-related objective function (\eg, cross-entropy for classification). The traditional paradigm adds a penalty term $P(\bm{K})$ by a pre-defined strength factor $\lambda$, 
\begin{equation}\label{eq-loss}
L_{\text{total}}(X, Y, \bm{\Theta}) = L_{\text{perf}}(X, Y, \bm{\Theta}) + \lambda P(\bm{K}) \,,
\end{equation}
where the common forms of $P$ include L1 \cite{li2016pruning}, L2 \cite{ding2018auto}, and group Lasso \cite{liu2015sparse,wen2016learning}. Specifically, group Lasso is effective in producing channel-wise structured sparsity. In the following discussions, we denote a specific channel in $\bm{K}$ by $\bm{F}^{(j)} = \bm{K}_{j,:,:,:}$. Then the group Lasso loss is formulated as 
\vskip -0.08in
\begin{equation}\label{eq-lasso}
P_{\text{Lasso}}(\bm{K}) = \sum_{j=1}^{D} ||\bm{F}^{(j)}||_E \,,
\end{equation}
\vskip -0.08in
where $||\bm{F}^{(j)}||_E$ is the Euclidean norm 
\vskip -0.08in
\begin{equation}
||\bm{F}||_E=\sqrt{\sum_{c=1}^{C} \sum_{p=1}^{K} \sum_{q=1}^{K} \bm{F}_{c,p,q} ^ 2} \,.
\end{equation}
\vskip -0.08in

With $G(\bm{F})$ as the gradient, we take the derivative,
\begin{equation}\label{eq-grad-lasso-W}
G(\bm{F}) = \frac{\partial L_{\text{total}}(X, Y, \bm{\Theta})}{\partial \bm{F}} = \frac{\partial L_{\text{perf}}(X, Y, \bm{\Theta})}{\partial \bm{F}} + \lambda \frac{\bm{F}}{||\bm{F}||_E} \,.
\end{equation}


The training dynamics of a specific channel $\bm{F}$ are quite straightforward. Beginning from a well-trained model, $\bm{F}$ resides near the local optima, thus the first term of Eq. \ref{eq-grad-lasso-W} is close to $\bm{0}$ but the second is not, so $\bm{F}$ is pushed closer to $\bm{0}$. If $\bm{F}$ is important to the performance, the objective function will intend to maintain its magnitude, \ie, the first gradient term will compete against the second, thus $\bm{F}$ will end up smaller than it used to be, depending on $\lambda$. Otherwise, taking the extreme case for example, if $\bm{F}$ does not influence $L_{\text{perf}}$ at all, the first term will be $\bm{0}$, so $\bm{F}$ will keep growing towards $\bm{0}$ by the second term. In other words, the performance-related loss and the penalty loss compete so that the resulting value of $\bm{F}$ will reflect its importance, which we refer to as \textit{competence-based importance evaluation} for convenience. However, we face a dilemma. \textbf{Problem A}: The penalty deviates the parameters of every channel from the optima of the objective function. Notably, a mild deviation may not bring negative effects, \eg, L2 regularization can also be viewed as a mild deviation. However, with a strong penalty, though some channels are zeroed out for pruning, the remaining channels are also made too small to maintain the representational capacity, which is an undesired side-effect. \textbf{Problem B}: With mild penalty for the high resistance, we cannot achieve high prunability, because most of the channels merely become closer to $\bm{0}$ than they used to be, but not close enough for perfect pruning.

We propose to achieve high prunability with a mild penalty by resetting the gradients derived from the objective function. We introduce a binary mask $m\in \{0, 1\}$, which indicates whether we wish to zero out $\bm{F}$. For the ease of implementation, we add no terms to the objective function (\ie, $L_{\text{total}} = L_{\text{perf}}$), simply derive the gradients as usual, and then manually apply the mask, add the penalty gradients and use the resultant gradients for SGD update:
\vskip -0.08in
\begin{equation}\label{eq-grad-lasso-W-m}
G(\bm{F}) \gets\frac{\partial L_{\text{perf}}(X, Y, \bm{\Theta})}{\partial \bm{F}}m + \lambda \frac{\bm{F}}{||\bm{F}||_E} \,.
\end{equation}
\vskip -0.08in
We will describe how to decide which channels to zero out (\ie, set mask values for multiple channels) in the next section. In this way, we have solved the above two problems. \textbf{A)} Though we add Lasso gradients to the objective-related gradients of every channel, which is equivalent to deviating the optima by adding Lasso loss to the original loss, the deviation is mild ($\lambda=10^{-4}$ in our experiments) hence harmless to the performance. \textbf{B)} With $m=0$, the first term no longer exists to compete against the second, thus even a mild $\lambda$ can make $\bm{F}$ steadily move towards $\bm{0}$.

\subsection{The Remembering Parts Remember Always, \\the Forgetting Parts Forget Progressively}


If directly used on conv kernels, \textit{Res} brings a problem: some objective-related gradients encode the supervision information for maintaining the performance but are discarded. Intuitively, the parameters are forced to ``forget'' some useful information (gradients). Fortunately, \textit{Rep is exactly the solution}, which allows us to prune the compactors only, not the original conv layers. ResRep only forces the compactors to ``forget'', and all the other layers still focus on ``remembering'', so we will not lose the information encoded in the gradients of the original kernels.

To combine Res with Rep, we need to decide which channels of $\bm{Q}$ to be zeroed out. When training the re-parameterized model, we add the Lasso gradients to the compactors only. After a few epochs, $||\bm{Q}_{j,:}||$ will reflect the importance of channel $j$ (\textit{competence-based importance evaluation} discussed in the Sect. \ref{sect-gr}), so we start to perform \textit{channel selection} based on the value of $\bm{Q}$. Let $n$ be the number of compactors, $\bm{m}^{(i)}$ (a $D^{(i)}$-dimensional binary vector) be the mask for the $i$-th compactor, we define $\bm{t}^{(i)}\in\mathbb{R}^{D^{(i)}}$ as the metric vector,
\begin{equation}
t^{(i)}_{j} = ||\bm{Q}^{(i)}_{j,:}||_2 \,, \quad \forall 1 \leq j \leq D^{(i)} \,.
\end{equation}
For each time of channel selection, we calculate the metric values for every channel in every compactor and organize them as a mapping $\mathcal{M} = \{(i, j) \to t^{(i)}_j \,|\, \forall 1\leq i \leq n, 1\leq j \leq D^{(i)}\}$. Then we sort the values of $\mathcal{M}$ in ascending order, start to pick one at a time from the smallest, and set the corresponding mask $m^{(i)}_j$ to 0. We stop picking when the reduced FLOPs \footnote{We count the theoretical FLOPs of the model without current mask-0 channels as current FLOPs. Reduced FLOPs = original - current FLOPs.} reaches our target, or we have already picked $\theta$ (named the \textit{channel selection limit}) channels. The mask values of unpicked channels are set to 1. The motivation is straightforward: following the discussions of competence-based importance evaluation, just like the traditional usage of penalty loss to compete against the original loss and select the channels with smaller norms, we use the penalty gradients to compete with the original gradients. Even better, all the metric values are 1 at the beginning (because every compactor kernel is initialized as an identity matrix), \textit{making it fair to compare them among different layers}. We initialize $\theta$ as a small number, increase $\theta$ every several iterations and re-select channels to ``forget'' progressively, avoiding zeroing out too many channels at once. As Fig. \ref{fig-comparison} shows, those mask-0 channels will become very close to $\bm{0}$, thus the structured sparsity emerges in compactors.

\section{Experiments}

\subsection{Datasets, Models and Settings}

We use ResNet-50 and MobileNet \cite{mobilev1} on \textbf{ImageNet-1K}. For the reproducibility, we follow the data augmentation of PyTorch official example~\cite{torch-example} including random cropping and flipping. For ResNet-50, we use the official torchvision base model (76.15\% top-1 accuracy) \cite{torch-model} for the fair comparison with most competitors. For MobileNet, we train from scratch with an initial learning rate of 0.1, batch size of 512 and cosine learning rate annealing for 70 epochs. The top-1 accuracy is 70.78\%, slightly higher than that reported in the original paper. We use ResNet-56/110 on \textbf{CIFAR-10}~\cite{krizhevsky2009learning} with the standard data augmentation \cite{he2016deep}: padding to $40\times40$, random cropping and flipping. We train the base models with batch size of 64 and the common learning rate schedule which is initialized as 0.1, multiplied by 0.1 at epoch 120 and 180, and terminated after 240 epochs. We count the FLOPs as multiply-adds, which is 4.09G for ResNet-50 \cite{torch-model}, 569M for MobileNet, and 126M/253M for ResNet-56/110.

\subsection{Pruning Results on ImageNet and CIFAR-10}

\begin{table*}
	\caption{Pruning results of ResNet-50 and MobileNet on ImageNet.}
	\label{exp-table-imgnet}
	\vspace{-0.2in}
	\begin{center}
		\begin{small}
			
			\begin{tabular}{llccccccc}
				\hline
				Model & Result 							&\makecell{Base\\ Top-1}	&\makecell{Base\\ Top-5}	&\makecell{Pruned\\ Top-1}	&\makecell{Pruned\\ Top-5}	& \makecell{Top-1 $\downarrow$ }	&	\makecell{Top-5 $\downarrow$ }	& 	\makecell{FLOPs $\downarrow$\%}	 			\\
				\hline
				\multirow{17}{*}{ResNet-50}	&	SFP \cite{he2018soft}	&	76.15 	&	92.87	&	74.61	&	92.06	&	1.54	&	0.81 	&	41.8	\\
				&GAL-0.5 \cite{GAL}				&	76.15	&	92.87	&	71.95	&	90.94	&	4.20		&	1.93		&	43.03	\\
				&NISP \cite{yu2018nisp}			&	-		&	-		&	-		&	-		&	0.89	&	-   	&	44.01	\\
				&Taylor-FO-BN \cite{molchanov2019importance}	&76.18	&	-	&74.50	& -		&	1.68	&	-		&	44.98	\\	
				&Channel Pr \cite{he2017channel}&	- 		&	92.2	&	-		&	90.8	&	- 		&	1.4 	&	50		\\
				&HP \cite{xu2018hybrid}			&	76.01 	&	92.93	&	74.87	&	92.43	&	1.14 	&	0.50 	&	50		\\
				&MetaPruning \cite{Metapruning}	&	76.6	&	-		&	75.4	&	-		&	1.2		&	-		&	51.10	\\
				&Autopr \cite{luo2018autopruner}&	76.15 	&	92.87	&	74.76	&	92.15	&	1.39	&	0.72 	&	51.21	\\
				&GDP \cite{lin2018accelerating}	&	75.13 	&	92.30	&	71.89	&	90.71	&	3.24 	&	1.59 	&	51.30	\\
				&FPGM \cite{FPGM}				&	76.15	&	92.87	&	74.83	&	92.32	&	1.32	&	0.55	&	53.5	\\
				&\textbf{ResRep}			&	\textbf{76.15}	&	\textbf{92.87}	&	\textbf{76.15}$\pm$\textbf{0.01}	&	\textbf{92.89}$\pm$\textbf{0.04}	&	\textbf{0.00}	&	\textbf{-0.02}	&	\textbf{54.54}	\\
				&C-SGD (extension) \cite{CSGD-journal}&	76.15	&	92.87	&	75.29	&	92.39	&	0.86 	&	0.48 	&	55.44	\\
				&DCP \cite{zhuang2018discrimination}&76.01 	&	92.93	&	74.95	&	92.32	&	1.06 	&	0.61 	&	55.76	\\		
				&C-SGD \cite{CSGD}				&	75.33	&	92.56	&	74.54	&	92.09	&	0.79 	&	0.47 	&	55.76	\\
				&ThiNet \cite{DBLP:journals/pami/LuoZZXWL19}&	75.30	&	92.20	&	72.03	&	90.99	&	3.27 	&	1.21 	&	55.83	\\
				&SASL \cite{SASL}				&	76.15	&	92.87	&	75.15	&	92.47	&	1.00	&	0.40	&	56.10	\\
				&\textbf{ResRep}			&	\textbf{76.15}	&	\textbf{92.87}	&	\textbf{75.97}$\pm$\textbf{0.02}	&	\textbf{92.75}$\pm$\textbf{0.01}	&	\textbf{0.18}	&	\textbf{0.12}	&	\textbf{56.11}	\\
				&TRP \cite{TRP}					&	75.90	&	92.70	&	72.69	&	91.41	&	3.21	&	1.29	&	56.52	\\
				&LFPC \cite{he2020learning}		&	76.15	&	92.87	&	74.46	&	92.32	&	1.69	&	0.55	&	60.8	\\
				&HRank \cite{HRank}				&	76.15	&	92.87	&	71.98	&	91.01	&	4.17	&	1.86	&	62.10	\\
				&\textbf{ResRep} 		&	\textbf{76.15}	&	\textbf{92.87}	&	\textbf{75.30}$\pm$\textbf{0.01}	&	\textbf{92.47}$\pm$\textbf{0.01}	&	\textbf{0.85}	&	\textbf{0.40}	&	\textbf{62.10}	\\
			
				\hline
				
				\multirow{2}{*}{MobileNet}	
				&MetaPruning \cite{Metapruning}			&	70.6	&	-		&	66.1	&	-		&	4.5		&	-		&	73.81		\\
				&\textbf{ResRep}	&	\textbf{70.78}	&	\textbf{89.78}	&	\textbf{68.02}$\pm$\textbf{0.02}	&	\textbf{87.66}$\pm$\textbf{0.02}	&	\textbf{2.76}	&	\textbf{2.12}	&	\textbf{73.91}		\\
				
				
				\hline
			\end{tabular}
		\end{small}
	\end{center}
	\vspace{-0.15in}
\end{table*}

We apply ResRep on ResNet-50 and MobileNet with the same hyper-parameters: $\lambda=10^{-4}$, batch size=256, initial learning rate=0.01 and cosine annealing for 180 epochs. We set the channel selection limit $\theta=4$ and $\theta\gets \theta+4$ every 200 batches and the first channel selection begins after 5 epochs. That is, after a 5-epoch ``warm-up'', we pick up 4 channels with the lowest $t$ values among all the layers and then pick 4 more channels every 200 batches, until we reach the FLOPs reduction target. For the ease of comparison, we experiment with ResNet-50 for three times with FLOPs reduction target of 54.5\% (1\% higher than FPGM \cite{FPGM}), 56.1\% (SASL \cite{SASL}) and 62.1\% (HRank \cite{HRank}), respectively, and MobileNet with 73.9\% to compare with MetaPruning \cite{Metapruning}. Following most competitors, we prune the first ($1\times1$) and second ($3\times3$) conv layers in every residual block of ResNet-50, and every non-depthwise conv of MobileNet. Inspired by a prior work \cite{GSM} which zeros out some gradients and utilizes momentum and weight decay for CNN sparsification, we raise the SGD momentum coefficient \textit{on compactors} from 0.9 (the default setting in most cases) to 0.99. Intuitively, the mask-0 channels continuously grow in the same direction (\ie, towards zero), and such a tendency accumulates in the momentum, thus the zeroing-out process can be accelerated by a larger momentum coefficient. For ResNet-56/110, the target layers include the first layers of residual block, and we use the same hyper-parameters as ImageNet except batch size of 64 and 480 training epochs. 
\begin{table}
	\caption{Pruning results of ResNet-56/110 on CIFAR-10.}
	\label{exp-table-cifar}
	\vspace{-0.2in}
	\begin{center}
		\begin{small}
			\setlength{\tabcolsep}{1.5mm}{
				\begin{tabular}{llcccc}
					\hline
					Model & Result 							& \makecell{Base \\ Top-1}		&\makecell{Pruned \\Top-1}	& \makecell{Top-1 \\ $\downarrow$\%} 	& 	\makecell{FLOPs \\ $\downarrow$\%}	 			\\
					\hline
					\multirow{7}{*}{R56}	
					&	AMC \cite{he2018amc}					&	92.8	&	91.9			&	0.9		&	50		\\
					&	FPGM \cite{FPGM}						&	93.59	&93.26	&	0.33	&	52.6\\
					&	SFP \cite{he2018soft}					&	93.59	&93.35	&	0.24	&	52.6\\
					&	LFPC \cite{he2020learning}				&	93.59	&93.24	&	0.35	&	52.9\\
					&	\textbf{ResRep}			&	\textbf{93.71}	&	\textbf{93.71$\pm$0.02}	&\textbf{0.00}	&\textbf{52.91}	\\	
					&	TRP \cite{TRP}							&	93.14	&	91.62			&	1.52	&	77.82	\\
					&	\textbf{ResRep}			&	\textbf{93.71}	&	\textbf{92.66$\pm$0.07}		&\textbf{1.05}	&	\textbf{77.83}	\\
					
					\hline
					
					\multirow{4}{*}{R110}	& Li et al. \cite{li2016pruning}		&	93.53	&	93.30	&	0.23 		&	38.60	 	\\
					&	GAL-0.5 \cite{GAL}						&	93.50	&	92.74	&	0.76		&	48.5		\\
					&	HRank \cite{HRank}						&	93.50	&	93.36	&	0.14		&	58.2		\\
					&	\textbf{ResRep}	&	\textbf{94.64}	&	\textbf{94.62$\pm$0.04}	&	\textbf{0.02}	&	\textbf{58.21}\\
					
					\hline
				\end{tabular}
			}
		\end{small}
	\end{center}
	\vspace{-0.25in}
\end{table}

Table. \ref{exp-table-imgnet}, \ref{exp-table-cifar} show the superiority of ResRep. Our results are average of 3 runs on ImageNet and 5 runs on CIFAR. On ResNet-50, ResRep achieves 0.00\% top-1 accuracy drop, which is the first to realize lossless pruning with such high pruning ratio (54.54\%), to the best of our knowledge. In terms of top-1 accuracy drop, ResRep outperforms SASL by 0.82\%, HRank by 3.32\% and all the other recent competitors by a large margin. On MobileNet, ResRep outperforms MetaPruning by 1.77\%. On ResNet-56/110, ResRep also outperforms, even though the comparison on accuracy drop is biased towards other methods, as our base models have higher accuracy (\ie, it is more challenging to prune a higher-accuracy model without accuracy degradation).

\begin{figure*}
	\begin{subfigure}{0.33\linewidth}
		\includegraphics[width=\linewidth]{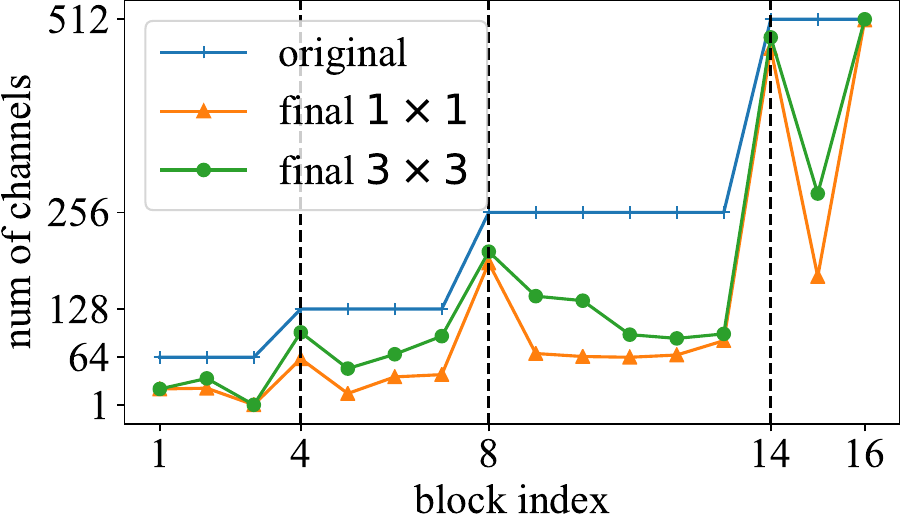} 
	\end{subfigure}
	\begin{subfigure}{0.33\linewidth}
		\includegraphics[width=\linewidth]{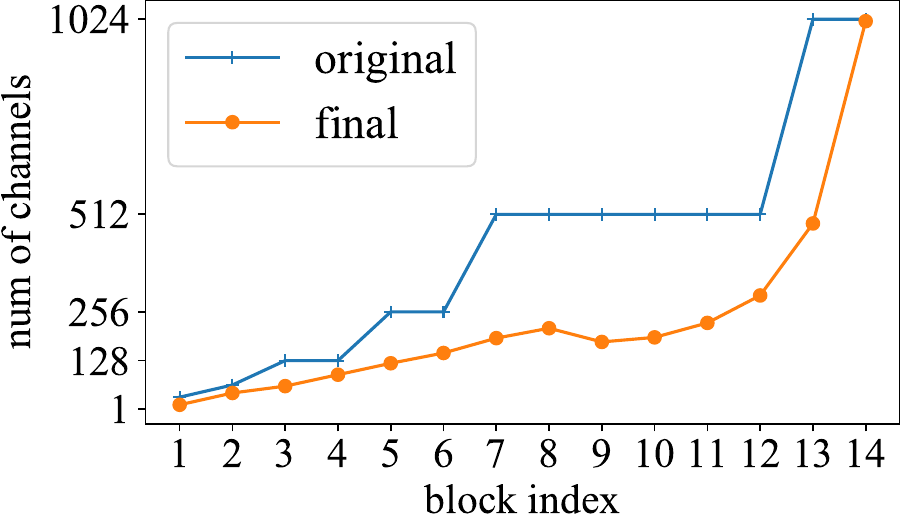} 
	\end{subfigure}
	\begin{subfigure}{0.33\linewidth}
		\includegraphics[width=\linewidth]{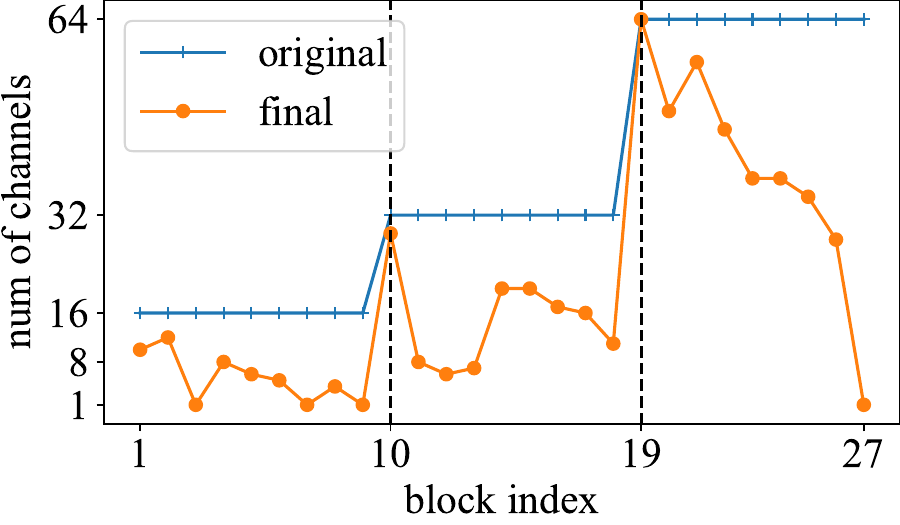} 
	\end{subfigure}
	\vspace{-0.1in}
	\caption{Width of target layers in pruned models. Left: ResNet-50 with the first $1\times1$ and $3\times3$ layer in each block shown separately. Middle: MobileNet. Right: ResNet-56. Vertical dashed lines indicate the stage transition in ResNets.}
	\label{fig-width}
	\vskip -0.1in
\end{figure*}

The final width of each target layer (Fig. \ref{fig-width}) shows that ResRep discovers the appropriate final structure without any prior knowledge, given the desired global pruning ratio. Notably, ResRep chooses to preserve more channels at higher-level layers of ResNet-50 and MobileNet, but prunes aggressively on the last blocks of ResNet-56. An explanation is that rich higher-level features are essential for maintaining the fitting capacity on difficult task like ImageNet, while ResNet-56 suffers from over-fitting on CIFAR-10. 

\begin{figure}
	\begin{subfigure}{0.47\linewidth}
		\includegraphics[width=\linewidth]{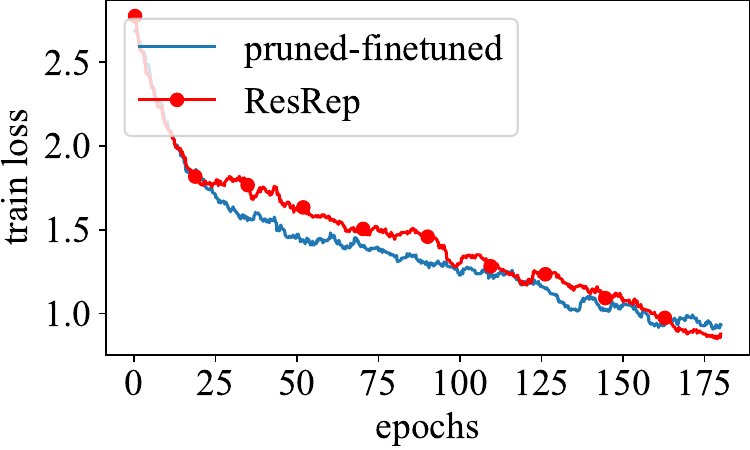} 
	\end{subfigure}
	\begin{subfigure}{0.47\linewidth}
		\includegraphics[width=\linewidth]{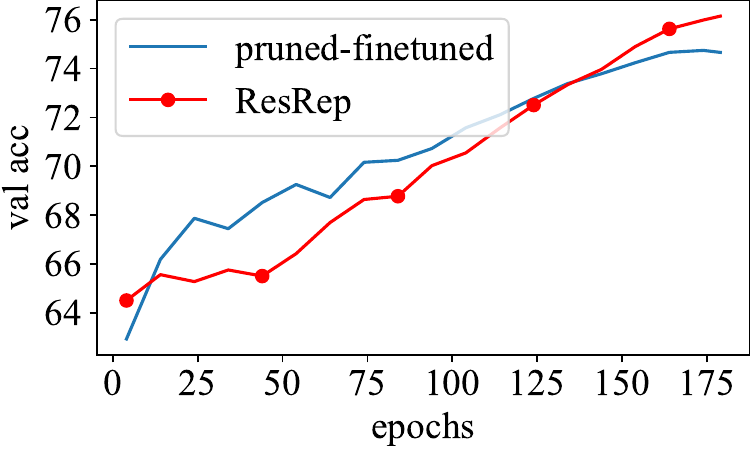} 
	\end{subfigure}
	\vspace{-0.1in}
	\caption{Training loss (left) and validation accuracy (right) of pruned-finetuned and ResRep on ResNet-50.}
	\label{fig-pruned-finetuned}
	\vskip -0.15in
\end{figure}

\begin{table}
	\caption{ResNet-50 baselines and variants of ResRep.}
	\label{exp-table-baselines}
	\vspace{-0.2in}
	\begin{center}
		\begin{small}
			\begin{tabular}{llcccc}
				\hline
				& Top-1 acc 	& 	FLOPs $\downarrow$\%	 \\
				\hline
				1) Base model finetuned			&	76.19		&	-		\\
				2) Uniformly shrunk baseline	&	74.39		&	55.4	\\
				3) Pruned-finetuned	&	74.66		&	54.5	\\
				\hline
				4) Vector re-parameterization		&	75.57		&	54.5	\\
				5) Momentum on compactors as 0.9		&	75.05		&	45.1	\\
				\hline
			\end{tabular}
		\end{small}
	\end{center}
	\vspace{-0.3in}
\end{table}

\begin{figure*}
	\begin{subfigure}{0.33\linewidth}
		\includegraphics[width=\linewidth]{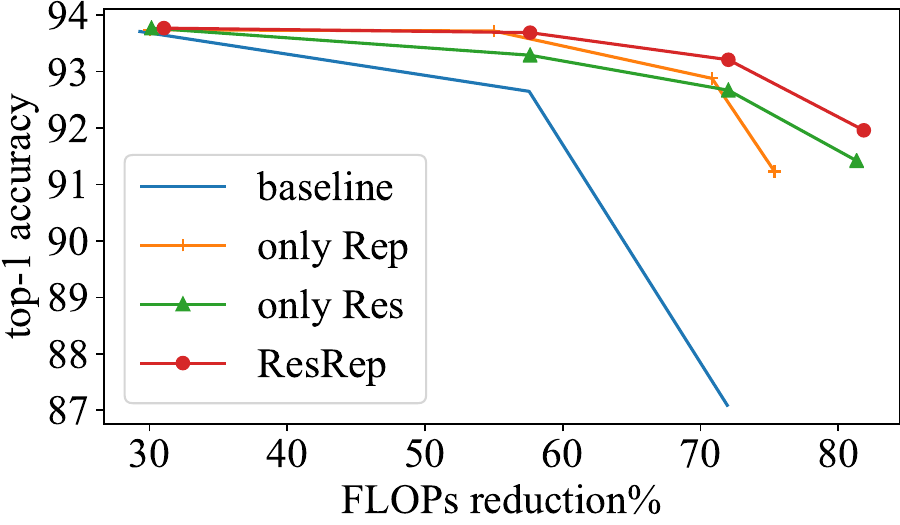} 
	\end{subfigure}
	\begin{subfigure}{0.33\linewidth}
		\includegraphics[width=\linewidth]{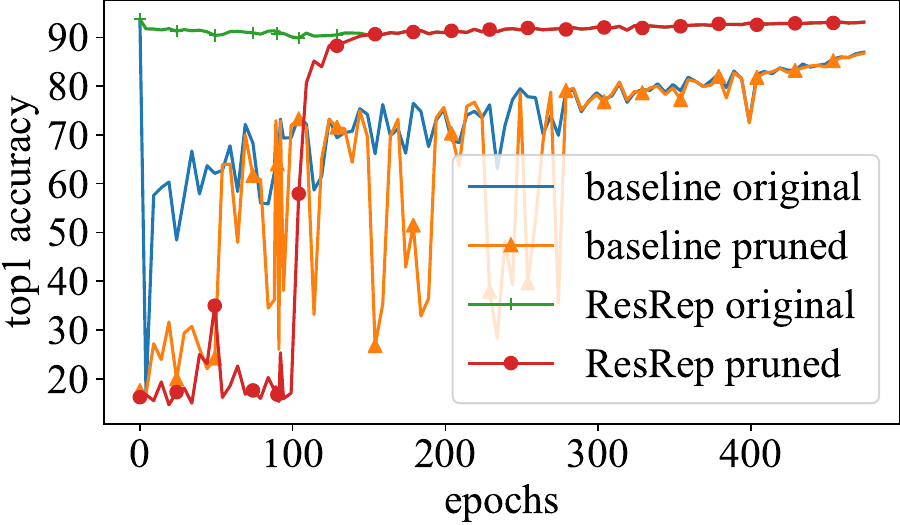} 
	\end{subfigure}
	\begin{subfigure}{0.33\linewidth}
		\includegraphics[width=\linewidth]{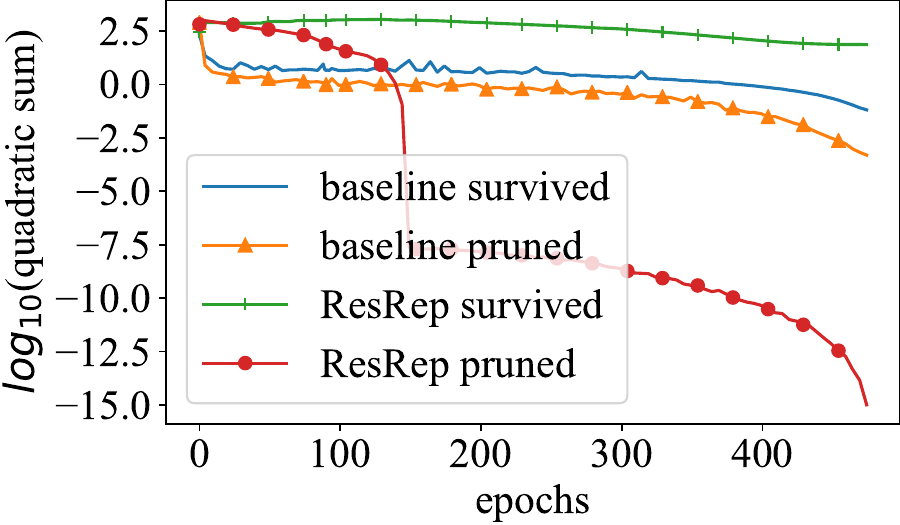} 
	\end{subfigure}
	\vspace{-0.1in}
	\caption{Left: FLOPs reduction v.s. accuracy of baseline, Res, Rep and ResRep. Middle: the original and pruned accuracy every 5 epochs. Right: the quadratic sum of survived parameters and those to-be-pruned (note the logarithmic scale).}
	\label{fig-comparison}
	\vskip -0.15in
\end{figure*}

We construct a series of baselines and variants of ResRep for comparison (Table. \ref{exp-table-baselines}). \textbf{1)} The base model is further finetuned with the same learning rate schedule for 180 epochs, and the accuracy is merely lifted by 0.04\%, suggesting the performance of ResRep is not simply due to the effect of training settings. \textbf{2)} We construct a uniformly shrunk baseline by building a ResNet-50 where all the target layers are reduced to 1/2 of the original width, thus the FLOPs is reduced by 55.4\%. We train it from scratch with the same training settings as the base model, and the final accuracy is 74.39\%, which is 1.58\% lower than our 56.1\%-pruned model. \textbf{3)} We compare ResRep against a classic pruning-then-finetuning method by sorting all the channels of base model by their Euclidean norm, pruning from the smallest until 54.5\% reduction ratio, and finetuning the resultant model with the same 180-epoch setting. Not surprisingly, the finetuned model is hardly better than the shrunk baseline trained from scratch. The training loss and validation accuracy (Fig. \ref{fig-pruned-finetuned}) shows the pruned model recovers quickly but cannot reach a comparable accuracy as the 54.5\%-pruned ResRep model. This observation is consistent with a prior study that a pruned model may be easily trapped into bad local minima \cite{liu2018rethinking}. \textbf{4)} A straightforward alternative of re-parameterization to verify the significance of Convolutional Re-parameterization. It changes the form of compactor from $1\times1$ conv to a $D$-dimensional trainable vector (\ie, a channel-wise scaling layer) initialized as $\bm{1}$, and the Lasso penalty naturally degrades to L1. After training with the same settings as the 54.5\%-pruned model for 180 epochs, the final accuracy is 75.57\%, which is 0.58\% lower. Intuitively, re-parameterization with $1\times1$ conv \textit{folds} the original kernel into a lower-dim kernel (\ie, re-combines the channels), but a vector simply \textit{deletes} some channels. \textbf{5)} We verify the necessity of 0.99-momentum by setting the momentum of compactors as 0.9, and fewer channels end up close to zero (below $\epsilon=10^{-5}$). Though a higher momentum reduces parameters faster, it is not necessary if longer training time is acceptable.


\subsection{Ablation Studies}\label{sect-ablation}

We then perform controlled experiments with the same training configurations as described above on ResNet-56 to evaluate Rep and Res separately. As the baseline, we adopt the traditional paradigm by directly adding Lasso loss (Eq. \ref{eq-lasso}) on all the target layers. With $\lambda\in\{0.3, 0.03, 0.003, 0.001\}$, we obtain four models with different final accuracy: 69.81\%, 87.09\%, 92.65\%, 93.69\%. To realize perfect pruning on each trained model, we attain the \textit{minimal structure} by removing the channels one at a time until the accuracy drops below the original (\ie, pruning any one more channel of the minimal structure will decrease the accuracy). Then we record the FLOPs reduction of the minimal structures: 81.24\%, 71.94\%, 57.56\%, 28.31\%. We test Rep but no Res by applying Lasso loss on the compactors with varying $\lambda$ to achieve comparable FLOPs reduction as baselines. And with Res but no Rep, we directly apply Gradient Resetting on the original conv kernels, targeting at the same FLOPs reduction as the four baseline models. Then we experiment with the full-featured ResRep. As shown in the left of Fig. \ref{fig-comparison} (the baseline data point of (81.24\%, 69.81\%) is ignored for better readability), Res and Rep deliver better final accuracy than the baselines, and perform even better when combined. 

We investigate into the training process by saving the parameters of the $\lambda=0.03$ baseline every 5 epochs. After training, we obtain the minimal structure, turn back to prune each saved model into the minimal structure, and report the accuracy before and after pruning. For ResRep, we do the same but on the compactors instead of the original conv layers. Fig. \ref{fig-comparison} (middle) shows that the baseline accuracy drops drastically because of the side-effects brought by strong Lasso, which implies low resistance. In contrast, the original accuracy of ResRep maintains on a high level. The pruning-caused damage (original accuracy minus pruned accuracy) is great for both the baseline and ResRep models at the beginning but reduces as the sparsity emerges. The pruned accuracy of baseline improves slowly and unsteadily due to the competence of two losses.

For each saved model, we also collect the quadratic sum of parameters which survive at last as well as the quadratic sum of those finally pruned, according to the final minimal structure. Fig. \ref{fig-comparison} (right, note the logarithmic scale) shows that the parameters of baseline soon become too small to maintain the performance, which explains the poor resistance. For ResRep, the magnitude of survived parameters decreases but maintains on a high level due to the mild penalty, and those to-be-pruned (\ie, mask-0) parameters drop steadily and soon become very close to zero, which explains the high resistance and high prunability.

\section{Conclusion}

The effectiveness of ResRep suggests that decomposing the traditional learning-based pruning into ``performance-oriented learning'' and ``pruning-oriented learning'' may be a promising research direction. As a successful application of Structural Re-parameterization, ResRep uses the methodology of \textit{constructing extra structures that can be converted back}, which enables to adopt some custom techniques (an update rule on the compactors only, in this case). Such a methodology may be useful in other research areas.

{\small
	\bibliographystyle{ieee_fullname}
	\bibliography{resrepbib}
}

\end{document}